\documentclass[conference]{IEEEtran}
\IEEEoverridecommandlockouts
\usepackage{cite}
\usepackage{amsmath,amssymb,amsfonts}
\usepackage{algorithmic}
\usepackage{graphicx}
\usepackage{textcomp}
\usepackage{xcolor}

\usepackage{multirow, makecell}
\usepackage[margin=2cm]{geometry}
\usepackage{adjustbox}
\usepackage{placeins}
\usepackage[hyphens]{url}
\usepackage[bookmarks=true, bookmarksnumbered=true, bookmarksopen=true, unicode=true, colorlinks=true,
  linkcolor=blue, citecolor=blue, filecolor=blue, urlcolor=blue, pdfstartview=FitH]{hyperref}
\usepackage{url}
\usepackage{array}
\newcolumntype{P}[1]{>{\centering\arraybackslash}p{#1}}
\newcolumntype{M}[1]{>{\centering\arraybackslash}m{#1}}

\def\BibTeX{{\rm B\kern-.05em{\sc i\kern-.025em b}\kern-.08em
    T\kern-.1667em\lower.7ex\hbox{E}\kern-.125emX}}

\begin{document}

\title{Semantic and Visual Similarities for Efficient Knowledge Transfer in CNN Training}

\author{\IEEEauthorblockN{Lucas Pascal}
\IEEEauthorblockA{\textit{Orkis$^{1}$, EURECOM$^{2}$} \\
$^{1}$Aix-en-Provence, France \\
lpascal@orkis.com\\
$^{2}$Sophia Antipolis, France \\
lucas.pascal@eurecom.fr}
\and
\IEEEauthorblockN{Xavier Bost}
\IEEEauthorblockA{\textit{Orkis} \\
Aix-en-Provence, France \\
xbost@orkis.com}
\and
\IEEEauthorblockN{Benoit Huet}
\IEEEauthorblockA{\textit{EURECOM} \\
Sophia Antipolis, France \\
huet@eurecom.fr}
}

% For IEEE version
% \IEEEpubid{\makebox[\columnwidth]{978-1-7281-4673-7/19/\$31.00~\copyright2019 IEEE \hfill} \hspace{\columnsep}\makebox[\columnwidth]{ }}

\maketitle

% For IEEE version
% \IEEEpubidadjcol

\begin{abstract}
In recent years, representation learning approaches have disrupted many multimedia computing tasks. Among those approaches, deep convolutional neural networks (CNNs) have notably reached human level expertise on some constrained image classification tasks. Nonetheless, training CNNs from scratch for new task or simply new data turns out to be complex and time-consuming. Recently, transfer learning has emerged as an effective methodology for adapting pre-trained CNNs to new data and classes, by only retraining the last classification layer. This paper focuses on improving this process, in order to better transfer knowledge between CNN architectures for faster trainings in the case of fine tuning for image classification. This is achieved by combining and transfering supplementary weights, based on similarity considerations between source and target classes. The study includes a comparison between semantic and content-based similarities, and highlights increased initial performances and training speed, along with superior long term performances when limited training samples are available.
\end{abstract}


\vspace{2mm}

\noindent \textcolor{red}{\textbf{Cite as:}\\L.~Pascal, X.~Bost, B.~Huet.2019. \href{http://cbmi2019.org/}{Semantic and Visual Similarities for Efficient Knowledge Transfer in CNN Training.}\\ doi: \href{https://doi.org/1010101.1010101}{https://doi.org/1010101.1010101}}

\vspace{2mm}

\noindent {\textbf{Note:} This work is an updated version of the official CBMI article}.

\section{Introduction}
\label{sec:intro}

With the emergence of large, public and thoroughly annotated datasets \cite{Russakovsky2015}, along with the ever increasing computing capabilities of GPUs, Deep Neural Networks, and especially CNNs, have rapidly revolutionized many computer vision tasks.
Such quantities of data allow to learn visual feature extractors whose relevance and discrimination power surpasses the best hand crafted ones \cite{Krizhevsky2012}, regardless of the problem complexity or the model size. However, the time and associated cost for creating such new huge datasets, and to make new models converge over these are still a huge bottleneck in real-world use cases, so that someone with limited resources cannot reasonably compete with companies running each of their models during weeks over hundreds of GPUs (or TPUs) and gigantic datasets.

Transfer learning is a recent and still evolving approach to address this issue. It consists in reusing a model developed for a task as a starting point for another related task.
This is based on the assumption that two related tasks require some common knowledge, so that some of the knowledge associated to a task could benefit another similar one.
In deep learning, this is performed by using some of the model's weights as an initialization for the training over the new task, while the usual practice is to initialize them randomly (\cite{Glorot2010}). In computer vision tasks, well trained CNN low level and mid-level layers generally detect basic shapes and textures, no matter the specificity of the task. They consequently transfer well between different computer vision problems, as shown in numerous publications (\cite{Hinton2006}\cite{Yosinski2014}\cite{Chu2016} \cite{Azizpour2016}\cite{Zamir2018}\cite{Tamaazousti2017}\cite{Wang2017}).

Transfer learning can then be distinguished in two types of applications : domain adaptation, aiming to adapt a pre-trained network to a new task, out of this work's scope, and fine-tuning, which consists in adapting the network to new target data, for the same task.
In the latter case, one can extend the transfer to the whole set of weights concerning the features extraction, and just replace the output fully connected layer with one shaped for his target classes. Almost all the knowledge required to perform the task is already present in the network, and can be aligned with the target data by a small training procedure (not necessary on the entire set of weights), lighter than the one required to train a model ``from scratch''. Note that fine-tuning a pre-trained CNN often leads to better performances, and requires fewer data than a network trained from scratch, as most of the knowledge required for the task is already present in it (\cite{Yosinski2014, Chu2016}).

Fine tuning has become common practice, allowing faster trainings on consequently smaller datasets, and giving the opportunity for researchers and companies to develop their own systems. However, there may still be a lack of efficiency in training a new fully connected layer from scratch, and transfer learning can once again fulfill it. To further improve the transfer, we thus propose to reuse some weights of the last fully connected layer of the original model, based on similarity between source and target classes.

The contribution of this work is fivefold: 
First, we show that there is some important knowledge within the last layer of pre-trained DNN models which when identified and used properly can be somewhat transferred to the new model, to speed up training and benefit model accuracy for fine tuning. 
Second, we propose a novel method to reuse that knowledge in combining multiple relevant source classes.
Third, we conduct a study over one visual and two semantic similarity measures to select these relevant source classes.
Fourth, we propose an original analysis enabling us to separate cases between three possible types of knowledge transfer, to attest that we are performing well on each of them, and optimize our process. 
Fifth, we monitor our method while decreasing the amount of training data, to validate the consistency of the results.

The paper is organized as follows. We will first review some works related to ours, then present our approach, before discussing the experimental results. Finally, we will present the conclusions of this work, and propose future developments.

\vspace{0.5cm}
\section{Related Works}

\subsection{Datasets and architectures}
The most common source dataset to apply transfer learning for computer vision tasks is ImageNet \cite{Russakovsky2015}, since it presents $1000$ classes, shared between various semantic fields (animals, flora life, vehicles, tools, etc...). It is thus very likely to benefit the training of almost any kind of target data, and its efficiency is demonstrated in \cite{Huh2016}. 

As for the choice of the network to use, many state of the art results in image classification tasks (including the ImageNet dataset) have been achieved by (or built on) the ResNet architecture \cite{He2016,He2016a}, and Inception \cite{Christian2015, Christian2017} structures (or combinations of them). Their efficiency and simplicity often places them as the best choice for transfer learning, be it for other classification tasks, or using it as a backbone for other tasks (object detection and segmentation, for example).

\vspace{0.2cm}
\noindent
\subsection{Transfer Learning process}

Reusing some pre-trained weights for a new task has been pioneered by \cite{Hinton2006} and \cite{Yosinski2014}, showing that the new task can greatly benefit it, not only in terms of training speed, but also of global performance. 

However, the way to optimize a transfer learning process is still unclear. We know that the deeper we go in a network, the more specialized are its weights to the task they are trained on \cite{Krizhevsky2012, Yosinski2014}. Concretely, if the first few layers of a ResNet (detecting simple visual patterns like geometrical shapes) can benefit any computer vision task, it is still unclear how deep the weights can efficiently be reused. \cite{Yosinski2014} experiments transfers of different depths, and highlights that the transfer of specialized layers can hurt performance on the target task, depending on their depth.

In \cite{Chu2016} is given a study taking into account the amount of data available. They show that if transferring weights has only a moderate impact on performance in a context with a lot of data, it becomes crucial for long-term performance as the data decreases. 
For two sufficiently close tasks (source and target), they generally advise to transfer all the layers except the classification one before a global fine tuning. This advice seems quite reasonable in the case we are investigating, since only the images contained in the dataset and their classes change, while the classification objective remains the same.

Hoewever, with enough populated classes, \cite{Yosinski2014} shows that training only the randomly initialized part of the transferred CNN can break fragile co-adaptations at the boundary. Fine tuning equally the whole network gives better results, allowing to readjust those co-adaptations. \cite{Tamaazousti2017} proposes a finer process that consists in training the whole network in one go with a lower learning rate applied to the transfered part. This focuses the training on the new part, while allowing co-adaptation between the two parts.

As shown in \cite{Wang2017}, transfer learning can also be improved by deepening and/or widening the original network, giving more rooms to small adjustments, under the condition of correctly managing the simultaneous training of both transferred and newly created cells.

A systematic process in all these works is to discard the classifying layer and to train a new one, adapted to the target classes, from scratch. We argue and show that, when the source and target are similar to a certain extent, the knowledge contained in the pre-trained classifier layer can be efficiently reused for learning a new model. 

\vspace{0.2cm}
\noindent
\subsection{Semantic similarity between textual content} 

One traditional way to get a similarity measure between two concepts is to use the WordNet graph \cite{Beckwith1991}: WordNet is an english lexical database of nouns, verbs, adjectives, and adverbs grouped under lexicalized concepts (named synsets), interlinked by different types of semantic relations.  There are five main semantic similarity measures defined for WordNet in the literature : Jiang \& Conrath \cite{Jiang1997}, Leacock \& Chodorow \cite{Leacock1998a}, Lin \cite{Lin1998}, Resnik \cite{Resnik1995}, and Wu \& Palmer \cite{Wu1994}. Each of them evaluates the semantic distance between two synsets. \cite{Capelle2012} evaluated the Wu \& Palmer one, making use of the path length between the synsets organized in a 'is-a' hierarchy and the depth of their most specific ancestor node, as the best one for semantic similarities.

More recently, \cite{Mikolov2013} designed an approach using neural networks to project words into feature vectors named Word2Vec representations, to represent efficiently textual content. In this feature space, distances between words are shown to be quite accurate to attest and quantify some semantic relationships \cite{Mikolov2013, Wang2014, Handler2014}.

\vspace{0.5cm}
\section{Similarity-based knowledge transfer}

A standard, basic transfer learning process to train an image classifier for some target classes is to reuse the convolution weights of a network already trained on a similar task on source classes. A fully connected layer fitting the target task is then randomly initialized on top of the network, which is fine tuned following a strategy adapted to the specificities of the problem (amount of available data, possible computational power restrictions, etc...). %!

The fully connected layer of the pre-trained network represents the knowledge of the task it is devised for. In cases where the original classification problem and the destination one are close, there might be a gain in transferring some of the knowledge of the original network head (last layer) to the target one. %!

We hypothesize that re-adjusting this available knowledge to fit the target classes could be more efficient than creating it from scratch, in the usual way. Assume we dispose of $M$ source classes and $N$ target ones, each target class (represented by its fully connected weights) could be initialized with a combination of a relevant subset of those $M$ source classes, instead of randomly. We define the relevance of such a subset of classes by using alternative similarity measures.

\vspace{0.2cm}
\subsection{Similarity measures}
We propose three of them. The first is a visual one, directly based on the image content. The two others are label-based semantic similarities :

\textbf{Inference similarity}. The images of the target classes are input to the plain source network. The similarities between source and target classes are computed as F-score measures for each "source network output/target class" couples. In a more practical way, let $o_{j}$ be the $j^{th}$ output of the source network and $c_{i}$ the $i^{th}$ target class with $j \in \{1, ..., M\}$ and $i \in \{1, ..., N\}$, we compute $sim(i,j)$ as the F-score for $o_{j}$ discriminating $c_{i}$. This similarity measure aims at leveraging relations based more on pure visual content than semantics.

\textbf{WordNet similarity}, using the Wu \& Palmer (\cite{Wu1994}) measure, as advised in \cite{Capelle2012}.
    
\textbf{Word2Vec similarity}. Using some pre-trained Word2Vec embeddings, we compute the standard cosine similarity between the Word2Vec embeddings of the source and target class names.

In the following section, these three initialization techniques are compared to the classic one (i.e. using random initialization of the neural network weights).

\vspace{0.2cm}
\subsection{Initialization}

We consider the affinity values as the coefficients of a neighboring structure, allowing us to approach the target class as a combination of some source neighbors. We thus compute, for each target class, the weights of its classifier as a linear combination of its $K$ closest source neighbors with respect to the similarity measure, taking as coefficients these affinity values (normalized, for them to sum to one over $K$).

\begin{eqnarray}
W'_{i} = \sum_{j=1}^{K} \bigg( \frac{sim(i,j)}{\sum_{j=1}^{K} sim(i,j)} \bigg) W_{j} \nonumber
\end{eqnarray}

In this way, each of the $K$ source classes neighbors contributes to the construction of the target class initialization, in proportion to its normalized similarity. The setting of $K$ with respect to the target classes is studied in the experimental part.

\vspace{0.5cm}
\section{Experiments and Results}

\subsection{Implementation Details}
In the following, to combine architecture simplicity and high performances, we use a ResNet-101 pre-trained on Imagenet, and replace the last fully connected layer to fit the target classes. We train weights from the fourth block (included) to the end, and freeze the rest. One could use a finer transfer strategy to optimize the results obtained \cite{Yosinski2014, Chu2016, Wang2016, Tamaazousti2017}.
Input images' smallest sides are resized to $256$ (preserving aspect ratio), then cropped (randomly for training, center crop for testing) to output $224\times224$ images. 

We used Adam optimizer with a learning rate of $10^{-3}$, $\beta_{1}=0.9$, $\beta_{2}=0.999$ and $\epsilon=10^{-8}$, with a batch size of 64. We apply a dropout of 0.75 on the last fully connected layer to prevent overfitting.

\vspace{0.2cm}
\subsection{Dataset}
For this experiment, we choose as source classes the $1000$ classes of the ILSVRC challenge \cite{Russakovsky2015}, which the ResNet-101 has been trained to classify. We take as target classes $90$ ImageNet synsets that are not part of those former $1000$. They can be separated in three types :

\textbf{Included classes}. The target synset is a child of a source synset, thus representing a more restrictive class than the one in the original problem.

\textbf{Inclusive classes}. The target class is an ancestor of some source synset(s), thus representing a more general class.

\textbf{Disjoint classes}. Neither child nor ancestor of any already source synset.

For these $90$ target classes we select some synsets containing at least $1000$ images and equally distributed into the three types of target classes ($30$ classes each). Within each target class, we pick $100$ images for testing and $900$ for training, producing balanced training and testing sets. Depending on the experiments, only a certain portion of this training set will be used for training. The list of synsets used for this experiment along with the selected images is available on a GitHub repository. \footnote{\url{https://github.com/lucaspascal/semantic-and-visual-similarities-for-efficient-knowledge-transfer-in-CNN-training}.}

\vspace{0.2cm}
\subsection{Similarities and Initialization}

We compute the inference similarities as explained earlier, with a pass of the training set through the pre-trained network (with the $1000$ class pre-trained classifier).

For the WordNet similarities, we use those given by the WordNet module of the NLTK Python library to obtain a similarity measure based on the shortest path that connects the labels (or synsets) in the "is-a" (hypernym/hyponym) taxonomy.

To compute the Word2Vec embedding of a given label, we average the embeddings \footnote{as trained on Flickr, and publicly available at \url{https://github.com/li-xirong/hierse/blob/master/README.md}.}  of all the words composing it, since labels are not always denoted by a single word, but often by an expression.
 
Each target class is then initialized with the weights of its selected source(s), depending on the chosen similarity measure and the number of source neighbors. For the random initialization baseline, we use a classic Xavier initialization \cite{Glorot2010}.

\vspace{0.2cm}
\subsection{Neighboring optimization}

\begin{figure}[h]
  \centering
  \includegraphics[width=\linewidth]{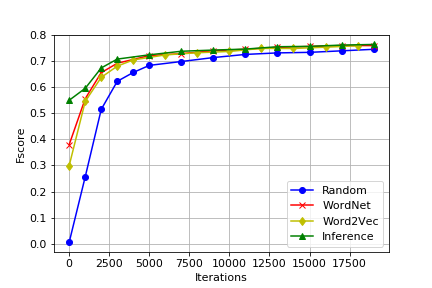}
  \caption{averaged per-class Fscore}
  \label{global fscore}
\end{figure}

We first show the global behavior of our initialization method with a single source class neighbor as initialization for each of the target class, in terms of F-score measure, averaged over the $90$ target classes. For this experiment, we populated each of the 90 target classes with $500$ training images (limit above which the results did not change significantly), and trained the networks over these.
\textbf{Fig.\ref{global fscore}} shows this evolution in terms of F-score averaged over all the 90 target classes, with the four initialization strategies (i.e Random, Inference, WordNet and Word2Vec).

We observe that each of the three initialization strategies performs better that the random baseline: the convergence is accelerated, and the models produce interesting
results even without training (iteration 0): from $40\%$ to more than $70\%$ of the F-score achieved at convergence, depending on the model. The initialization by inference similarity is performing best, as one could have expected 
 since the similarities in this case have directly been evaluated with respect to the task's performance metric. The four models tend to converge to the same value, provided with enough data to fill the gap.

% For IEEE VERSION
% We give in \textbf{Table~\ref{class_bullet}} an example of source/target class correspondence given by  each similarity measure, for the "\textit{Bullet, slug}" target class. The results of these transfers are shown in \textbf{Fig.~\ref{single classes}}. In this case, the Word2Vec method has been mistaken by the "\textit{slug}" term, and chose the mollusc as a source class (worst performing). WordNet found a logical source class ("\textit{Projectile, missile}"), according to the synsets semantic, and  the inference similarity selected "\textit{Lipstick, lip rouge}", which has no obvious semantic link with a bullet, but presents some very similar visual patterns, as shown on the images (performing best).

\begin{figure}[h!]
  \centering
  \includegraphics[width=\linewidth]{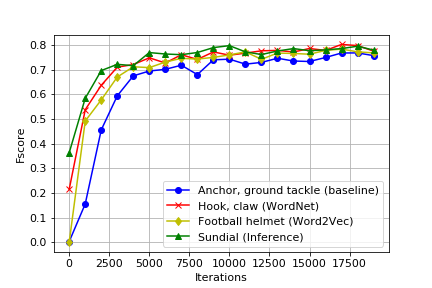}
  \caption{Evolution of the models on the target class \textit{Anchor, ground tackle}, for the different source classes determined by the similarity measures.}
  \label{single classes}
\end{figure}

% For HAL VERSION
We give in \textbf{Table~\ref{class_anchor}} an example of source/target class correspondence given by  each similarity measure, for the "\textit{Anchor, ground tackle}" target class. The results of these transfers are shown in \textbf{Fig.~\ref{single classes}}. In this case, the Word2Vec method has been mistaken by the "\textit{tackle}" term, and chose the football accessory as a source class (worst performing). WordNet found a logical source class ("\textit{Hook, claw}"), according to the synsets semantic, and  the inference similarity selected "\textit{Sundial}", which has no obvious semantic link with an anchor, but presents some very similar visual patterns, as shown on the images (performing best).

From this example, along with the global results, we conclude that label-based semantic similarities are more likely to select wrong matchings for visual classification,  while the inference similarity is able to bring out better ones, out of any semantic consideration.

% For HAL VERSION
Another intersting fact here is that the Word2Vec similarity is still learning faster than the random initialization. This observation, verified over multiple other examples, suggests that any pre-trained classifier is always a better initialization than a random one for fine-tuning in image classification.

% For IEEE VERSION
% \begin{table}[h!]
% \begin{center}
% \caption{Classes correspondances} \label{tab:cap}
% \begin{tabular}{|*{4}{M{1.5cm}|}}
%   \hline
%   % after \\: \hline or \cline{col1-col2} \cline{col3-col4} ...
%   \textbf{Target Class} & \textbf{Inference Affinity} & \textbf{WordNet Affinity} & \textbf{Word2Vec Affinity}
%   \\
%     \hline
%     \hline
%     bullet, slug & lipstick, lip rouge & projectile, missile & slug \\
%     \hline 
%     \includegraphics[width=0.8\linewidth]{bullet_1.JPEG}    \includegraphics[width=0.8\linewidth]{bullet_3.JPEG} &
%   \includegraphics[width=0.66\linewidth]{lipstick_2.JPEG}  \includegraphics[width=0.55\linewidth]{lipstick_3.JPEG} &
%   \includegraphics[width=0.67\linewidth]{projectile_2.JPEG}  \includegraphics[width=0.67\linewidth]{projectile_3.JPEG} &
%     \includegraphics[width=0.8\linewidth]{slug_2.JPEG}  \includegraphics[width=0.8\linewidth]{slug_3.JPEG}\\
%     \hline
% \end{tabular}
% \label{class_bullet}
% \end{center}
% \end{table}

% For HAL VERSION
\begin{table}[h!]
\begin{center}
\caption{Classes correspondances} \label{tab:cap}
\begin{tabular}{|*{4}{M{1.5cm}|}}
  \hline
  \textbf{Target Class} & \textbf{Inference Affinity} & \textbf{WordNet Affinity} & \textbf{Word2Vec Affinity}
  \\
    \hline
    \hline
    Anchor, ground tackle & Sundial & Hook, claw & Football helmet \\
    \hline 
    \includegraphics[width=0.9\linewidth]{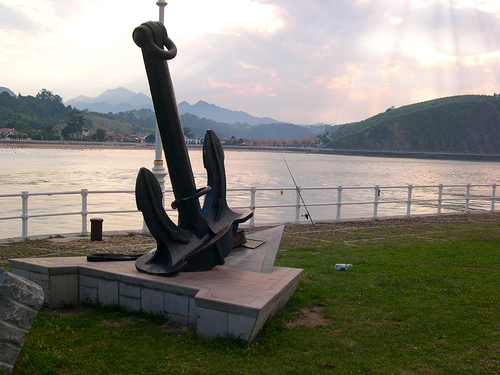}
    \includegraphics[width=0.9\linewidth]{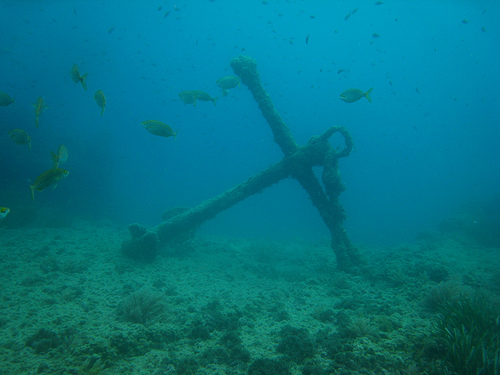} &
  \includegraphics[width=0.9\linewidth]{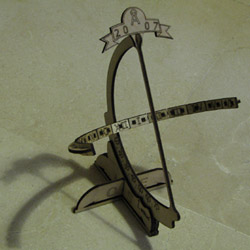}
  \includegraphics[width=0.9\linewidth]{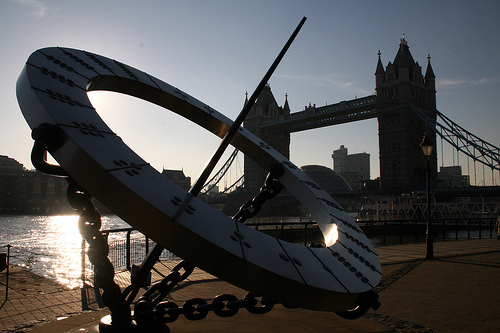} &
  \includegraphics[width=0.4\linewidth]{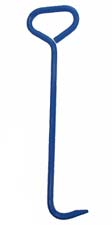} 
  \includegraphics[width=0.6\linewidth]{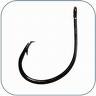} &
    \includegraphics[width=0.8\linewidth]{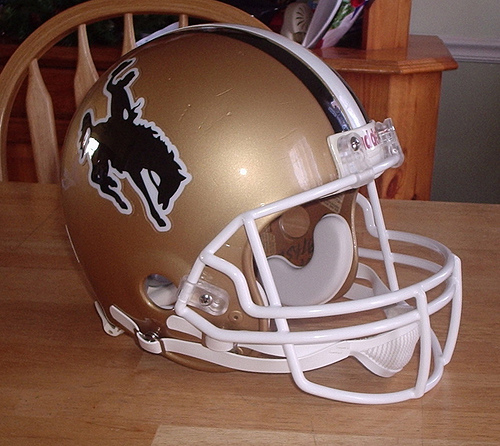}
    \includegraphics[width=0.8\linewidth]{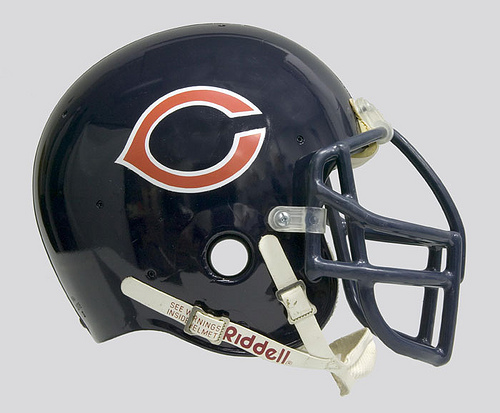}\\
    \hline
\end{tabular}
\label{class_anchor}
\end{center}
\end{table}

% For IEEE VERSION
% \begin{figure}[h!]
%   \centering
%   \includegraphics[width=\linewidth]{bullet_no_title.png}
%   \caption{Evolution of the models on the target class \textit{Bullet, slug}, for the different source classes determined by the similarity measures.}
%   \label{single classes}
% \end{figure}

% For HAL VERSION

\begin{figure*}[h!]
  \centering
  \includegraphics[width=0.32\linewidth]{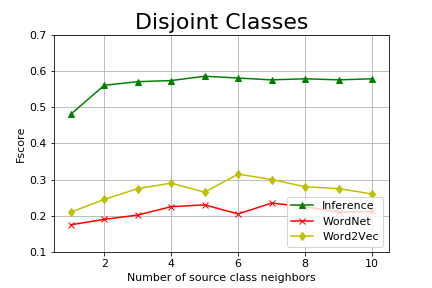}
  \includegraphics[width=0.32\linewidth]{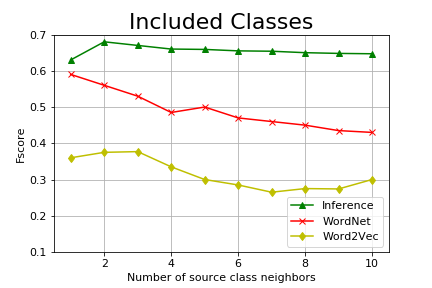}
  \includegraphics[width=0.32\linewidth]{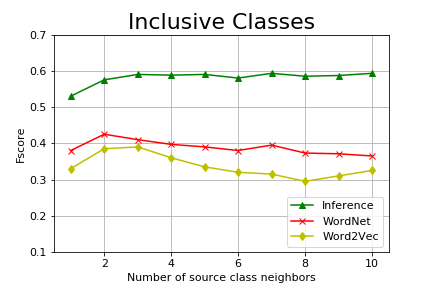}
  \caption{Immediate inference results for each type of classes and initialization, with respect to the number of source classes selected to compute the initialization.}
  \label{knn init}
\end{figure*}

We then extend the study to the use of multiple source classes neighbors to compute the different initializations:
\textbf{Fig.~\ref{knn init}} shows the F-scores of the models directly after initialization (without training), with respect to the number of source class neighbors selected, for each type of target class (i.e disjoint, included and inclusive). 

% \begin{figure}[h!]
%   \centering
%   \includegraphics[width=\linewidth]{images/k_disjoint_samescale.png}
%   \includegraphics[width=\linewidth]{images/k_included_samescale.png}
%   \includegraphics[width=\linewidth]{images/k_inclusive_samescale.png}
%   \caption{Immediate inference results for each type of classes and initialization, with respect to the number of source classes selected to compute the initialization.}
%   \label{knn init}
% \end{figure}

\begin{table*}[h!]
\begin{center}
\caption{Evolution of the four methods through data reduction. Performances after initialization (left columns) and best registered performances (right columns) are reported, with respect to the number of training samples for each class.} \label{tab:cap}
\begin{tabular}{|*{9}{M{1.5cm}|}}
  \hline
\multirowcell{2}{\textbf{Images} \\ \textbf{per class}} & \multicolumn{2}{M{3.2cm}|}{\textbf{Random}}  & \multicolumn{2}{M{3.2cm}|}{\textbf{Visual similarity}} & \multicolumn{2}{M{3.2cm}|}{\textbf{WordNet semantic similarity}} & \multicolumn{2}{M{3.2cm}|}{\textbf{Word2Vec semantic similarity}} \\\cline{2-9}

  & \textbf{First} & \textbf{Best} & \textbf{First} & \textbf{Best} & \textbf{First} & \textbf{Best} & \textbf{First} & \textbf{Best} \\
    % \hline
    % 500 & 0.0067 & 0.7724 & \textbf{0.6019} & \textbf{0.7913} & 0.3907 & 0.7902 & 0.3593 & 0.7876 \\
    \hline
    100 & 0.00 & 0.72 & \textbf{0.59} & \textbf{0.73} & 0.39 & 0.72 & 0.35 & 0.72 \\
    \hline
    50 & 0.00 & 0.68 & \textbf{0.58} & \textbf{0.69} & 0.39 & 0.68 & 0.35 & 0.68 \\
    \hline
    25 & 0.00 & 0.62 & \textbf{0.58} & \textbf{0.64} & 0.39 & 0.63 & 0.35 & 0.63 \\
    \hline
    10 & 0.00 & 0.53 & \textbf{0.54} & \textbf{0.54} & 0.39 & 0.54 & 0.35 & 0.53 \\
    \hline
    5 & 0.00 & 0.41 & \textbf{0.50} & \textbf{0.50} & 0.39 & 0.43 & 0.35 & 0.45 \\
    \hline
    2 & 0.00 & 0.26 & \textbf{0.44} & \textbf{0.44} & 0.39 & 0.39 & 0.35 & 0.35 \\
    \hline
    1 & 0.00 & 0.16 & \textbf{0.40} & \textbf{0.40} & 0.39 & 0.39 & 0.35 & 0.35 \\
    \hline
\end{tabular}
\label{tableau}
\end{center}
\end{table*}

The initialization by inference similarity benefits the most from extending the number of source class neighbors: for any type of initialization, the built classifiers smoothly gain in performances by adding neighbors. Beyond the selection of one best source class, this confirms the superiority of the visual similarity over the semantic ones to estimate the relevance of any source class for a transfer.
For the WordNet and Word2Vec cases, there is also a significant gain, even if the evolution over the number of neighbors is more chaotic, and it appears to be a good way to compensate bad matchings (like the Word2Vec case in \textbf{Table~\ref{class_bullet}}).

\vspace{0.2cm}
\subsection{Data reduction study} \label{datareduction}
We then study how well this process generalizes while decreasing the amount of training data. For each of the initialization methods, we initialize a new model, taking for each target class the optimal number of neighbors source classes depending on its type (disjoint, included or inclusive). These optimal numbers are taken from \textbf{Fig.\ref{knn init}}.
\textbf{Table~\ref{tableau}} shows the scores of these models compared to the random baseline for $100$, $50$, $25$, $10$, $5$, $2$ and $1$ training images per class. For each model, the initial performance after initialization (without training) and the best registered performance until convergence  are reported.

The source classes selection by inference similarity varies with the number of training images (unlike the two others), since it is computed with those images. Its performance at initialization thus decreases with the amount of training data. However, it still always achieves better performances, which puts aside the idea of combining both types of similarities \cite{Safadi2014}. Under $5$ training images per class, a consequent performance gap remains between the baseline and our models even after training. Under $2$ images per class ($5$ for inference similarity), the best scores are achieved right after initialization, and training only degrades performances. Building the best possible initialization is thus crucial in such cases.

\vspace{0.5cm}
\section{Conclusion and Perspectives}
This paper addresses transfer learning in an image classification context. In particular, we proposed to study alternative approaches to re-use the knowledge inherent within the original pre-trained deep network in the target one (handling new image classes). Rather than only transferring network weights corresponding to the feature extraction part, we investigated several initialization strategies to re-use and combine specifically identified weights from the pre-trained classifier into the target model. To validate the impact of our method, we presented and tested three different similarity estimators, one visual and two semantics, optimized the models across the different types of target classes, and monitored them while reducing the amount of data.

In the end, our method produced systematically better initializations, faster trainings, and significantly superior long term performances in limited training data configurations. The consistency with which the best model, based on visual similarities, outperforms the baseline across the different types of target classes and amounts of data, along with its computational lightness (a few supplementary inferences in the network) suggest that it can be systematically adopted when performing transfer learning in this context.

\bibliographystyle{IEEEbib}
\bibliography{smart_transfer_arranged}

\end{document}